\documentclass[journal]{IEEEtran}
\usepackage{amsmath,amsfonts}
\usepackage{algorithmic}
\usepackage{algorithm}
\usepackage{array}
\usepackage[caption=false,font=normalsize,labelfont=sf,textfont=sf]{subfig}
\usepackage{textcomp}
\usepackage{stfloats}
\usepackage{url}
\usepackage{verbatim}
\usepackage{graphicx}
\usepackage{cite}
\usepackage{multirow}
\usepackage[table]{xcolor}
\usepackage{booktabs}

\begin{document}

\title{LUCID: \textbf{L}earned \textbf{U}ndersampling-Adaptive \textbf{C}onsistency-Guided \textbf{I}nference with \textbf{D}eterministic Flow Matching for Sparse-View CT Reconstruction}

% \author{Jigang Duan, Jiayi Wang, Heran Wang, Ping Yang, Genwei Ma$^\star$,~\IEEEmembership{Member,~IEEE}, and Xing Zhao$^\star$
\author{Jigang Duan, Jiayi Wang, Heran Wang, Ping Yang, Genwei Ma$^\star$, and Xing Zhao$^\star$
\thanks{This work was supported by the National Natural Science Foundation of China (Grant No. 12426308) and the Beijing High Innovation Plan ("Capital High-End Leading Talents Aggregation and Cultivation Program"), Grant No. 202504841094. \textit{(Jigang Duan and
Jiayi Wang contributed equally to this work.)} \textit{(Corresponding authors: Genwei Ma and Xing Zhao.)}}
\thanks{Jigang Duan, Jiayi Wang, Ping Yang, Heran Wang, and Xing Zhao are with the School of Mathematical Sciences, Capital Normal University, Beijing 100048, China (e-mail: hiduanjigang@163.com; wangjy152022@126.com; q1203479069@163.com; yang\_ping0603@163.com; zhaoxing\_1999@126.com).}
\thanks{Genwei Ma is with the National Center for Applied Mathematics Beijing, Capital Normal University and Academy for multidisciplinary studies, Capital Normal University, Beijing, 100048, China (e-mail: magenwei@126.com).}
\thanks{This work has been submitted to the IEEE for possible publication.
Copyright may be transferred without notice, after which this version may no longer be accessible.}}

% The paper headers
\markboth{Journal of \LaTeX\ Class Files,~Vol.~14, No.~8, August~2021}%
{Shell \MakeLowercase{\textit{et al.}}: A Sample Article Using IEEEtran.cls for IEEE Journals}

% \IEEEpubid{0000--0000/00\$00.00~\copyright~2021 IEEE}
% Remember, if you use this you must call \IEEEpubidadjcol in the second
% column for its text to clear the IEEEpubid mark.

\maketitle

\begin{abstract}
  Sparse-view CT reduces radiation dose and scanning time by acquiring fewer projection views, but angular undersampling makes reconstruction severely ill-posed, causing streak artifacts, structural blurring, and loss of fine details. Existing supervised methods are often tied to specific sampling settings, whereas generative methods may introduce anatomically inconsistent hallucination-like structures under severe undersampling. We propose Lucid, a sparsity-adaptive, consistency-guided reconstruction framework based on a Flow Matching generative prior for sparse-view CT. Lucid is trained only on high-quality CT images to learn a continuous transport between a Gaussian distribution and the high-quality CT image distribution, independent of view sampling. During inference, the sampling sparsity level is explicitly incorporated to adapt the generative trajectory of a single pretrained model. Specifically, Lucid constructs a degradation-matched initial state by sparsity-weighted fusion of the sparse-view FBP image and Gaussian noise, performs sparsity-modulated Flow Matching updates, and applies projection-domain data-consistency correction after each prior update. Experiments under multiple sparse-view settings show that Lucid achieves stable reconstruction performance across different sampling densities, improves image quality and structural fidelity, and reduces the risk of hallucination-like structures in generative sparse-view CT reconstruction.
\end{abstract}


\section{Introduction}

\IEEEPARstart{C}{omputed} tomography (CT) has been widely used in medical diagnosis, industrial nondestructive testing, and security screening because of its noninvasive acquisition, high spatial resolution, and capability of depicting anatomical and material structures. Conventional high-quality CT imaging relies on dense angular sampling over a full scan range, which inevitably increases radiation exposure and acquisition time. Sparse-view CT (SVCT) reduces the number of projection views to enable low-dose and accelerated imaging, and has therefore received increasing attention. However, angular undersampling substantially worsens the ill-posedness of CT reconstruction, leading to severe streak artifacts, structural blurring, and loss of fine details. Stable and faithful reconstruction from sparse-view projections thus remains a central challenge in CT imaging~\cite{wang2020deep,wang2024review,koetzier2023deep}.

\subsection{Trends of Models and Algorithms}

The objective of CT reconstruction is to recover the spatial distribution of the linear attenuation coefficient from X-ray projections. Let \(\mu(\boldsymbol{r})\) denote the underlying image, where \(\boldsymbol{r}\in\mathbb{R}^{2}\) is the spatial coordinate. For an ideal two-dimensional parallel-beam geometry, the projection model is given by the Radon transform
\begin{equation}
p(\theta,s)
=
\mathcal{R}\mu(\theta,s)
=
\int_{\mathbb{R}^{2}}
\mu(\boldsymbol{r})
\delta(s-\boldsymbol{r}\cdot\boldsymbol{n}_{\theta})
d\boldsymbol{r},
\label{eq:radon_transform}
\end{equation}
where \(\theta\) is the projection angle, \(s\) is the detector coordinate, \(\boldsymbol{n}_{\theta}=(\cos\theta,\sin\theta)^{T}\), and \(\delta(\cdot)\) denotes the Dirac delta function. After discretization, the CT forward model can be written as
\begin{equation}
\boldsymbol{y}
=
\boldsymbol{A}\boldsymbol{x}
+
\boldsymbol{\epsilon},
\label{eq:ct_forward_model}
\end{equation}
where \(\boldsymbol{x}\in\mathbb{R}^{N}\) is the image vector, \(\boldsymbol{y}\in\mathbb{R}^{M}\) is the measured projection data, \(\boldsymbol{A}\in\mathbb{R}^{M\times N}\) is the system matrix, and \(\boldsymbol{\epsilon}\) denotes noise and model mismatch.

In SVCT, only a subset of projection views is acquired. Let \(\Omega\) denote the sampled view set and \(\boldsymbol{P}_{\Omega}\) the corresponding view-selection operator. The sparse-view measurement model is
\begin{equation}
\boldsymbol{y}_{\Omega}
=
\boldsymbol{P}_{\Omega}\boldsymbol{A}\boldsymbol{x}
+
\boldsymbol{\epsilon}_{\Omega}.
\label{eq:sparse_view_model}
\end{equation}
Since \(|\Omega|\) is much smaller than the number of fully sampled views, directly solving Eq.~\eqref{eq:sparse_view_model} is severely ill-posed. Analytic reconstruction methods such as filtered back-projection (FBP)~\cite{ramachandran1971three} therefore often suffer from pronounced streak artifacts, structural blurring, and loss of fine details.

Model-driven methods mitigate this ill-posedness by incorporating explicit image priors into the physical imaging model, leading to the regularized reconstruction problem
\begin{equation}
\hat{\boldsymbol{x}}
=
\arg\min_{\boldsymbol{x}}
\frac{1}{2}
\left\|
\boldsymbol{P}_{\Omega}\boldsymbol{A}\boldsymbol{x}
-
\boldsymbol{y}_{\Omega}
\right\|_{2}^{2}
+
\lambda \mathcal{R}(\boldsymbol{x}),
\label{eq:regularized_reconstruction}
\end{equation}
where the data-fidelity term enforces consistency with the measured projections, \(\mathcal{R}(\boldsymbol{x})\) denotes the regularization prior, and \(\lambda\) balances the two terms. Total variation (TV) and its variants are representative choices~\cite{sidky2008image,lohvithee2017parameter,niu2017iterative,wang2018adaptive,xi2023adaptive}. They suppress streak artifacts by promoting gradient sparsity, but may cause staircase artifacts, edge degradation, and loss of fine structures. More advanced priors, including nonlocal regularization, dictionary learning, adaptive weighting, anatomical sparsity, dynamic filtering, and residual-guided strategies, have been developed to improve structural preservation~\cite{li2019few,zhao2019iterative,hu2019ordered,wang2021helical,shu2025nonlocal,yu2018dynamic,zhang2024robust}. Despite their interpretability and data consistency, these methods still rely heavily on handcrafted priors and parameter tuning, which limits their ability to represent complex anatomical structures under severe angular undersampling.

Deep learning methods have achieved substantial progress by learning reconstruction mappings from data. Image-domain methods directly map artifact-contaminated reconstructions to clean images~\cite{jin2017deep,zhang2018sparse,han2018framing}, whereas projection-domain methods complete or correct sparsely sampled sinograms~\cite{dong2019sinogram,lee2019deep}. To exploit complementary information, dual-domain methods jointly use projection-domain measurement constraints and image-domain structural priors~\cite{yuan2018sipid,lee2019high,hu2021hybrid,wu2021drone,shi2022dual,li2022ddptransformer}. Recent studies have further explored unified networks for multiple sampling settings to improve cross-view generalization~\cite{shi2025prompting,lin2026deepsparse}. However, supervised models are often tied to the sampling patterns, geometries, noise levels, and data distributions used during training. Their performance may therefore degrade when applied to different scanners, acquisition protocols, or sampling densities. Although unsupervised and self-supervised methods reduce the dependence on paired training data~\cite{kim2023sparsier2sparse,wu2024linear,xie2022limited}, stable recovery of fine structures remains challenging under severe undersampling.

Generative models provide another promising direction for SVCT reconstruction. Score-based generative models~\cite{song2019generative}, generative adversarial networks~\cite{goodfellow2020generative}, and diffusion models~\cite{ho2020denoising} have been used to introduce learned image priors into ill-posed reconstruction problems~\cite{lahiri2023sparse,chung2022diffusion,he2024solving,xu2024stage,yang2025ctsdm,liu2024sparse,wu2024multi}. These methods can effectively suppress sparse-view artifacts, but may also generate visually plausible yet anatomically inconsistent structures under out-of-distribution conditions, such as changes in view sampling or reconstruction settings~\cite{bhadra2021hallucinations,wu2024multi}. In medical CT, such hallucination-like structures may mimic lesions or obscure subtle abnormalities. Therefore, an effective generative reconstruction method should not only improve image quality, but also maintain consistency with the measured projections across different sampling densities.

Flow Matching (FM) is a generative modeling framework that learns a continuous transport between a source distribution and a target distribution through a time-dependent velocity field~\cite{lipman2022flow}. Let \(p_{0}\) denote the distribution of high-quality CT images and \(p_{1}\) denote a standard Gaussian distribution. Given \(\boldsymbol{x}_{0}\sim p_{0}\) and \(\boldsymbol{x}_{1}\sim p_{1}\), a simple linear path can be defined as
\begin{equation}
\boldsymbol{x}_{t}
=
(1-t)\boldsymbol{x}_{0}
+
t\boldsymbol{x}_{1},
\quad
t\in[0,1],
\label{eq:fm_path}
\end{equation}
where \(t=0\) corresponds to the clean image state and \(t=1\) corresponds to the noise state. The associated velocity field is
\begin{equation}
\boldsymbol{u}_{t}
=
\frac{d\boldsymbol{x}_{t}}{dt}
=
\boldsymbol{x}_{1}-\boldsymbol{x}_{0}.
\label{eq:true_velocity}
\end{equation}
FM trains a neural network \(\boldsymbol{v}_{\boldsymbol{\theta}}(\boldsymbol{x}_{t},t)\) to approximate this velocity field by minimizing
\begin{equation}
\min_{\boldsymbol{\theta}}
\mathbb{E}_{t,\boldsymbol{x}_{0},\boldsymbol{x}_{1}}
\left[
\left\|
\boldsymbol{v}_{\boldsymbol{\theta}}(\boldsymbol{x}_{t},t)
-
\boldsymbol{u}_{t}
\right\|_{2}^{2}
\right],
\label{eq:fm_loss}
\end{equation}
where \(t\) is sampled from \([0,1]\).

After training, samples are generated by solving the deterministic ordinary differential equation
\begin{equation}
\frac{d\boldsymbol{x}_{t}}{dt}
=
\boldsymbol{v}_{\boldsymbol{\theta}}(\boldsymbol{x}_{t},t),
\quad
\boldsymbol{x}_{t_{0}}=\boldsymbol{x}_{\mathrm{init}},
\quad
t:t_{0}\rightarrow 0.
\label{eq:fm_reverse_ode}
\end{equation}
A numerical discretization gives
\begin{equation}
\boldsymbol{x}_{t_{k+1}}
=
\boldsymbol{x}_{t_{k}}
+
(t_{k+1}-t_{k})
\boldsymbol{v}_{\boldsymbol{\theta}}(\boldsymbol{x}_{t_{k}},t_{k}),
\quad
t_{k+1}<t_{k},
\quad
t_{K}=0.
\label{eq:fm_discretization}
\end{equation}
Unlike diffusion models, FM does not require repeated stochastic noise injection during sampling. Its deterministic trajectory provides a controllable inference process, making it suitable for adapting the starting state and step-size schedule according to the current sampling sparsity. This property motivates the proposed sparsity-adaptive Flow Matching reconstruction framework.

\subsection{Motivation}

\begin{figure}[t]
\centering
\includegraphics[width=0.48\textwidth]{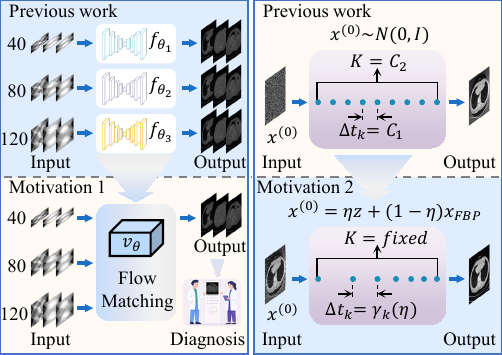}
\caption{Motivation of the proposed Lucid framework. Motivation 1 contrasts training-stage adaptation in existing supervised methods with inference-time adaptation using a unified Flow Matching prior. Motivation 2 contrasts fixed generative inference with the proposed sparsity-guided initialization and adaptive update scheduling.}
\label{fig:motivation}
\end{figure}

The degradation severity in sparse-view CT is closely related to the number of acquired projection views. Let \(\Theta\) denote the full-view angle set and \(\Omega\subset\Theta\) denote the sparse-view angle set. We define the sampling sparsity level as
\begin{equation}
\eta
=
1-\frac{|\Omega|}{|\Theta|},
\label{eq:sparsity_degree}
\end{equation}
where a larger \(\eta\) indicates more severe missing-view information. As \(\eta\) increases, the sparse-view FBP reconstruction usually contains stronger streak artifacts and more pronounced structural loss. Therefore, reconstructions under different sampling densities should not follow the same inference trajectory. Mildly undersampled data preserve more reliable anatomical structures and require conservative generative updates, whereas severely undersampled data need stronger prior guidance to suppress artifacts and recover missing structures.

As illustrated in Fig.~\ref{fig:motivation}, most supervised SVCT methods adapt to different sampling conditions at the training stage, either by training view-specific models~\cite{hu2021hybrid,wu2021drone,li2022ddptransformer,lahiri2023sparse} or by training a unified network on data mixed from multiple sampling rates~\cite{shi2025prompting,yang2025ctsdm,lin2026deepsparse}. Although these strategies improve applicability across predefined view settings, they do not explicitly adjust the inference trajectory according to the sparsity level of the current input. This limitation is more critical for generative reconstruction, where a mismatch between the degradation severity and the sampling trajectory may either leave severe artifacts insufficiently corrected or introduce unnecessary structural modifications for mildly degraded images.

The Flow Matching model used in this work is not trained as a supervised mapping from sparse-view artifacts to clean images. Instead, it learns a continuous transport path between a Gaussian source distribution and the distribution of high-quality CT images. Let \(\boldsymbol{X}_{0}\sim p_{0}\) denote a high-quality CT image and \(\boldsymbol{Z}\sim\mathcal{N}(\boldsymbol{0},\boldsymbol{I})\) denote a Gaussian source sample. The linear probability path used for training is
\begin{equation}
\boldsymbol{X}_{t}
=
(1-t)\boldsymbol{X}_{0}
+
t\boldsymbol{Z},
\quad
t\in[0,1].
\label{eq:motivation_fm_statistical_path}
\end{equation}
Under the mean-squared error objective, the optimal velocity field is
\begin{equation}
\boldsymbol{v}^{*}(\boldsymbol{x},t)
=
\mathbb{E}
\left[
\boldsymbol{Z}
-
\boldsymbol{X}_{0}
\mid
\boldsymbol{X}_{t}=\boldsymbol{x}
\right].
\label{eq:motivation_optimal_velocity}
\end{equation}
Thus, the pretrained network \(\boldsymbol{v}_{\boldsymbol{\theta}}(\boldsymbol{x},t)\) estimates the average forward transport direction along the learned probability path. During reverse-time inference, the negative velocity direction is used to move an intermediate state toward the high-quality CT image distribution. This provides a sampling-independent image prior, but its effectiveness during reconstruction depends on whether the inference starts from a state compatible with the learned probability path.

A pure Gaussian initialization is consistent with the Flow Matching source distribution but discards the structural information contained in the measured projections. In contrast, directly using the sparse-view FBP image preserves measurement-dependent structures but also introduces deterministic streak artifacts into the generative trajectory. To balance these two effects, we construct a sparsity-guided initial state as
\begin{equation}
\boldsymbol{x}^{(0)}
=
\eta\boldsymbol{z}
+
(1-\eta)\boldsymbol{x}_{\mathrm{FBP}},
\quad
\boldsymbol{z}\sim\mathcal{N}(\boldsymbol{0},\boldsymbol{I}).
\label{eq:motivation_sparsity_init}
\end{equation}
Let \(\boldsymbol{x}^{*}\) denote the underlying high-quality image and \(\boldsymbol{e}_{\Omega}\) denote the sparse-view FBP error:
\begin{equation}
\boldsymbol{x}_{\mathrm{FBP}}
=
\boldsymbol{x}^{*}
+
\boldsymbol{e}_{\Omega}.
\label{eq:motivation_fbp_decomposition}
\end{equation}
Substituting Eq.~\eqref{eq:motivation_fbp_decomposition} into Eq.~\eqref{eq:motivation_sparsity_init} gives
\begin{equation}
\boldsymbol{x}^{(0)}
=
\underbrace{
(1-\eta)\boldsymbol{x}^{*}
+
\eta\boldsymbol{z}
}_{\boldsymbol{x}_{\eta}^{\mathrm{FM}}}
+
(1-\eta)\boldsymbol{e}_{\Omega},
\label{eq:motivation_init_decomposition}
\end{equation}
where \(\boldsymbol{x}_{\eta}^{\mathrm{FM}}\) has the same form as an intermediate state on the Flow Matching training path at \(t=\eta\). Therefore,
\begin{equation}
\boldsymbol{x}^{(0)}
-
\boldsymbol{x}_{\eta}^{\mathrm{FM}}
=
(1-\eta)\boldsymbol{e}_{\Omega}.
\label{eq:motivation_init_bias}
\end{equation}
This indicates that the proposed initialization does not simply perturb the FBP image with noise. Instead, it embeds the sparse-view input near the learned Flow Matching path, with a residual bias determined by the FBP artifact term. When \(\eta\) is small, more reliable information from \(\boldsymbol{x}_{\mathrm{FBP}}\) is retained. When \(\eta\) is large, the initialization moves closer to the Gaussian source distribution, reducing the influence of severe FBP artifacts on the subsequent generative trajectory.

The same sparsity level also provides a natural way to regulate the reverse ODE integration. With a fixed number of inference steps \(K\), we adapt the reverse update step size according to
\begin{equation}
\Delta t_{k}
=
\gamma_{k}(\eta),
\quad
k=0,1,\ldots,K-1,
\label{eq:adaptive_schedule}
\end{equation}
where \(\Delta t_k>0\) denotes the effective reverse update step size. Conservative updates are preferred for mildly sparse data to avoid unnecessary modification of reliable structures, whereas stronger prior-driven updates are needed for severely sparse data to suppress pronounced artifacts. In this way, the generative trajectory is adjusted according to the actual sampling density rather than being fixed for all sparse-view inputs.

Finally, the generative prior should be constrained by the measured projection data. Lucid therefore alternates the sparsity-guided Flow Matching prior update with a projection-domain data-consistency correction. The prior update guides the reconstruction toward the learned high-quality CT image distribution, while the data-consistency step restricts the solution using the acquired sparse-view projections. Their combination allows a single pretrained Flow Matching prior to adapt to different sampling densities while reducing the risk of hallucination-like structures in generative sparse-view CT reconstruction.

\subsection{Our Contribution}

The main contributions of this work are summarized as follows:
\begin{itemize}
\item We propose Lucid, a sparsity-adaptive generative reconstruction framework for sparse-view CT based on deterministic Flow Matching. By learning a single image prior from high-quality CT images, Lucid enables inference-time adaptation without view-specific retraining.

\item We introduce a sampling sparsity-guided inference strategy that explicitly incorporates the sparsity level into the generative trajectory. Specifically, the sparsity level is used to construct a degradation-matched initial state and to modulate the Flow Matching update step size, allowing the same pretrained prior to adapt to different degrees of angular undersampling.

\item We integrate projection-domain data consistency into the Flow Matching inference process. After each prior-guided update, a data-consistency correction constrains the reconstruction with the measured sparse-view projections, improving structural fidelity and reducing hallucination-like local structures in generative sparse-view CT reconstruction.
\end{itemize}

\section{Method}
\label{sec:method}

\begin{figure*}[t] 
\centering 
\includegraphics[width=\textwidth]{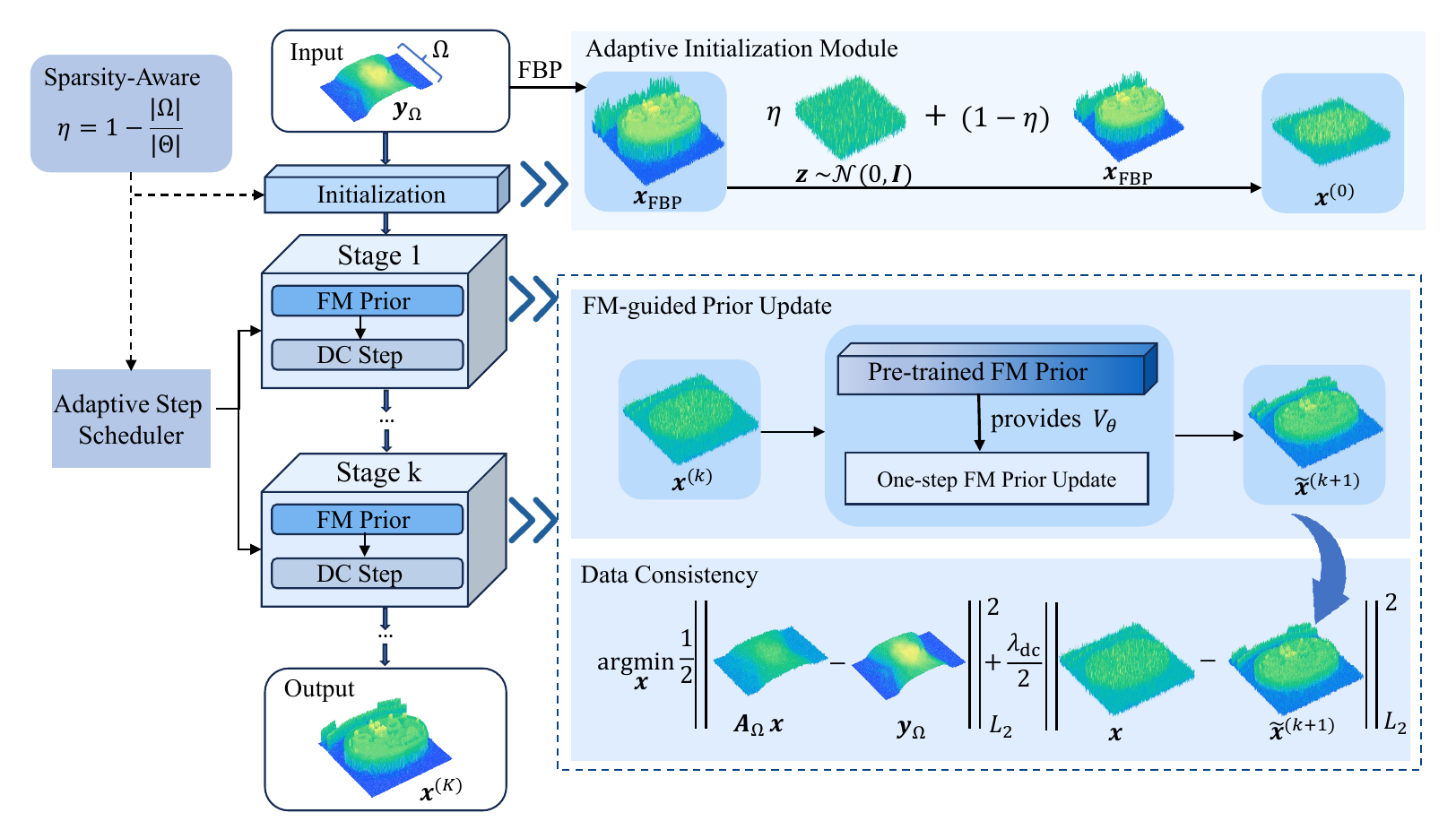} 
\caption{Overview of the proposed Lucid framework for sparse-view CT reconstruction. The Flow Matching prior is pretrained using only high-quality CT images. During inference, the sampling sparsity level is computed from the current view set, and a sparsity-guided initial state is constructed by combining the sparse-view FBP reconstruction with Gaussian noise. The reconstruction is then iteratively updated by alternating an FM-guided prior update with a projection-domain data-consistency step.}
\label{fig:lucid_framework} 
\end{figure*}

This section presents Lucid, a sparsity-adaptive Flow Matching framework for sparse-view CT reconstruction. As illustrated in Fig.~\ref{fig:lucid_framework}, Lucid uses the sampling sparsity level to determine both the initial state and the reverse-time step-size schedule during inference. Each Flow Matching prior update is followed by a projection-domain data-consistency step, so that the reconstruction is guided by the learned high-quality CT image prior while remaining consistent with the measured sparse-view projections.

\subsection{Overview of the Proposed Reconstruction Framework}

Let \(\Theta\) denote the full-view angle set and \(\Omega\subset\Theta\) denote the sampled sparse-view angle set. Following the sparse-view CT measurement model introduced in the Introduction, the sparse-view projection data are modeled as
\begin{equation}
    \boldsymbol{y}_{\Omega}
    =
    \boldsymbol{A}_{\Omega}\boldsymbol{x}
    +
    \boldsymbol{\epsilon}_{\Omega},
    \quad
    \boldsymbol{A}_{\Omega}
    =
    \boldsymbol{P}_{\Omega}\boldsymbol{A},
    \label{eq:method_sparse_model}
\end{equation}
where \(\boldsymbol{A}\) is the full-view system matrix, \(\boldsymbol{P}_{\Omega}\) is the view-selection operator, \(\boldsymbol{A}_{\Omega}\) is the sparse-view system matrix, and \(\boldsymbol{\epsilon}_{\Omega}\) denotes noise and model mismatch. We use the sampling sparsity level \(\eta\) defined in Eq.~\eqref{eq:sparsity_degree} to quantify the degree of angular undersampling.

Lucid employs a pretrained Flow Matching network \(\boldsymbol{v}_{\boldsymbol{\theta}}(\boldsymbol{x},t)\) as a generative prior for high-quality CT images. Instead of using a fixed inference trajectory, Lucid adapts both the initialization and the update step size according to \(\eta\). During inference, the reconstruction alternates between an FM-guided prior update and a projection-domain data-consistency correction:
\begin{equation}
    \boldsymbol{x}^{(k)}
    \xrightarrow{\ \mathrm{FM\ prior}\ }
    \tilde{\boldsymbol{x}}^{(k+1)}
    \xrightarrow{\ \mathrm{Data\ consistency}\ }
    \boldsymbol{x}^{(k+1)} ,
    \label{eq:overall_alternating}
\end{equation}
where \(\tilde{\boldsymbol{x}}^{(k+1)}\) denotes the intermediate image after the FM prior update, and \(\boldsymbol{x}^{(k+1)}\) denotes the data-consistency-corrected image. The following subsections detail the sparsity-guided initialization, adaptive update scheduling, and data-consistency correction.

\subsection{Flow Matching Prior for CT Image Reconstruction}

We use a pretrained Flow Matching model as a generative prior for high-quality CT images. Let \(\boldsymbol{x}_{0}\sim p_{0}\) denote a high-quality CT image sample, and let \(\boldsymbol{x}_{1}\sim\mathcal{N}(\boldsymbol{0},\boldsymbol{I})\) denote a Gaussian source sample. Following the definition in the Introduction, Flow Matching constructs a continuous path between the image distribution and the source distribution:
\begin{equation}
    \boldsymbol{x}_{t}
    =
    (1-t)\boldsymbol{x}_{0}
    +
    t\boldsymbol{x}_{1},
    \quad
    t\in[0,1],
    \label{eq:method_fm_path}
\end{equation}
where \(t=0\) corresponds to the high-quality image state and \(t=1\) corresponds to the Gaussian source state. The path velocity is given by
\begin{equation}
    \boldsymbol{u}_{t}
    =
    \frac{d\boldsymbol{x}_{t}}{dt}
    =
    \boldsymbol{x}_{1}
    -
    \boldsymbol{x}_{0}.
    \label{eq:method_true_velocity}
\end{equation}
The velocity network \(\boldsymbol{v}_{\boldsymbol{\theta}}(\boldsymbol{x}_{t},t)\) is trained to approximate this velocity field by minimizing
\begin{equation}
    \mathcal{L}_{\mathrm{FM}}
    =
    \mathbb{E}_{t,\boldsymbol{x}_{0},\boldsymbol{x}_{1}}
    \left[
    \left\|
    \boldsymbol{v}_{\boldsymbol{\theta}}(\boldsymbol{x}_{t},t)
    -
    \boldsymbol{u}_{t}
    \right\|_{2}^{2}
    \right].
    \label{eq:method_fm_loss}
\end{equation}

After training, the learned velocity field can be used for reverse-time inference from a noisy or intermediate state toward the image state. Given the current estimate \(\boldsymbol{x}^{(k)}\) and the time condition \(t_k\), one Euler-type Flow Matching prior update is written as
\begin{equation}
    \tilde{\boldsymbol{x}}^{(k+1)}
    =
    \boldsymbol{x}^{(k)}
    -
    \Delta t_k
    \boldsymbol{v}_{\boldsymbol{\theta}}
    \left(
    \boldsymbol{x}^{(k)}, t_k
    \right),
    \label{eq:method_fm_update_basic}
\end{equation}
where \(\Delta t_k>0\) denotes the effective reverse update step size. In Lucid, \(t_k\) is used as the time condition of the velocity network, while \(\Delta t_k\) controls the strength of the reverse prior update and is adaptively modulated according to the sampling sparsity level.

In Lucid, Flow Matching is not used as a standalone generator or an image-domain post-processing module. Instead, it serves as a learned image-distribution prior within the iterative reconstruction process. Each FM prior update is followed by a projection-domain data-consistency step, which constrains the generative update using the measured sparse-view projections and helps suppress hallucination-like structures inconsistent with the measured data.

\subsection{Sampling Sparsity-Guided Inference}

Lucid uses the sampling sparsity level \(\eta\), defined in Eq.~\eqref{eq:sparsity_degree}, to adapt two key components of the Flow Matching inference process: the initial state and the effective reverse update step size. In this way, different sparse-view settings are handled by trajectory adaptation during inference rather than by view-specific retraining.

\subsubsection{Sparsity-Guided Initialization}

Given the sparse-view projection data \(\boldsymbol{y}_{\Omega}\), we first compute the corresponding FBP reconstruction:
\begin{equation}
\boldsymbol{x}_{\mathrm{FBP}}
=
\operatorname{FBP}_{\Omega}
\left(
\boldsymbol{y}_{\Omega}
\right),
\label{eq:method_fbp}
\end{equation}
where \(\operatorname{FBP}_{\Omega}(\cdot)\) denotes the FBP operator under the current sparse-view geometry. To combine the measurement-dependent structure in \(\boldsymbol{x}_{\mathrm{FBP}}\) with the Gaussian source prior of Flow Matching, the initial state is constructed as
\begin{equation}
    \boldsymbol{x}^{(0)}
    =
    \eta \boldsymbol{z}
    +
    (1-\eta)\boldsymbol{x}_{\mathrm{FBP}},
    \quad
    \boldsymbol{z}
    \sim
    \mathcal{N}(\boldsymbol{0},\boldsymbol{I}).
    \label{eq:method_sparsity_init}
\end{equation}
Thus, more FBP information is retained for mildly undersampled data, whereas severely undersampled data are initialized closer to the Gaussian source distribution to reduce the influence of strong FBP artifacts.

\subsubsection{Sparsity-Guided Adaptive Step Size}

The sparsity level is also used to regulate the update strength of Flow Matching inference. With \(K\) prior-update steps, we assign the time condition at the \(k\)-th step as
\begin{equation}
    t_k
    =
    \eta
    \left(
    1
    -
    \frac{k}{K}
    \right),
    \quad
    k=0,1,\ldots,K-1.
    \label{eq:method_time_grid}
\end{equation}
This makes the reverse trajectory start from a sparsity-dependent time position and gradually approach the clean-image state.

We further define the sparsity modulation factor as
\begin{equation}
    g(\eta)
    =
    \frac{1+\alpha\eta}{1+\alpha},
    \label{eq:method_sparsity_modulation}
\end{equation}
where \(\alpha>0\) controls the modulation strength. The effective reverse update step size is then given by
\begin{equation}
    \Delta t_k
    =
    \Delta t_{\min}
    +
    \left(
    \Delta t_{\max}
    -
    \Delta t_{\min}
    \right)
    t_k^{\xi}
    g(\eta),
    \label{eq:method_adaptive_stepsize}
\end{equation}
where \(\Delta t_{\min}\) and \(\Delta t_{\max}\) are the lower and upper step-size bounds, and \(\xi>0\) controls the temporal decay rate. The factor \(t_k^{\xi}\) encourages a coarse-to-fine evolution, while \(g(\eta)\) increases the update strength for more severe angular undersampling.

\subsection{Projection-Domain Data Consistency Correction}

The Flow Matching prior update guides the current estimate toward the learned high-quality CT image prior, but it does not explicitly enforce consistency with the measured sparse-view projections. To constrain the generative update by the physical imaging model, we introduce a projection-domain data-consistency correction after each prior update.

Given the current estimate \(\boldsymbol{x}^{(k)}\), we first compute the adaptive step size using Eq.~\eqref{eq:method_adaptive_stepsize} and perform one Flow Matching prior update:
\begin{equation}
    \tilde{\boldsymbol{x}}^{(k+1)}
    =
    \boldsymbol{x}^{(k)}
    -
    \Delta t_k
    \boldsymbol{v}_{\boldsymbol{\theta}}
    \left(
    \boldsymbol{x}^{(k)}, t_k
    \right).
    \label{eq:method_fm_update}
\end{equation}
The intermediate image \(\tilde{\boldsymbol{x}}^{(k+1)}\) is then corrected by solving
\begin{equation}
\begin{aligned}
    \boldsymbol{x}^{(k+1)}
    =
    \arg\min_{\boldsymbol{x}}
    \frac{1}{2}
    \left\|
    \boldsymbol{A}_{\Omega}\boldsymbol{x}
    -
    \boldsymbol{y}_{\Omega}
    \right\|_{2}^{2}
    +
    \frac{\lambda_{\mathrm{dc}}}{2}
    \left\|
    \boldsymbol{x}
    -
    \tilde{\boldsymbol{x}}^{(k+1)}
    \right\|_{2}^{2},
\end{aligned}
\label{eq:method_dc}
\end{equation}
where \(\lambda_{\mathrm{dc}}>0\) controls the trade-off between projection-domain data fidelity and proximity to the FM prior update.

The corresponding normal equation is
\begin{equation}
    \left(
    \boldsymbol{A}_{\Omega}^{\top}
    \boldsymbol{A}_{\Omega}
    +
    \lambda_{\mathrm{dc}}\boldsymbol{I}
    \right)
    \boldsymbol{x}^{(k+1)}
    =
    \boldsymbol{A}_{\Omega}^{\top}
    \boldsymbol{y}_{\Omega}
    +
    \lambda_{\mathrm{dc}}
    \tilde{\boldsymbol{x}}^{(k+1)}.
    \label{eq:method_normal_equation}
\end{equation}
Equivalently, the formal solution can be written as
\begin{equation}
    \boldsymbol{x}^{(k+1)}
    =
    \left(
    \boldsymbol{A}_{\Omega}^{\top}
    \boldsymbol{A}_{\Omega}
    +
    \lambda_{\mathrm{dc}}\boldsymbol{I}
    \right)^{-1}
    \left(
    \boldsymbol{A}_{\Omega}^{\top}
    \boldsymbol{y}_{\Omega}
    +
    \lambda_{\mathrm{dc}}
    \tilde{\boldsymbol{x}}^{(k+1)}
    \right).
    \label{eq:method_dc_solution}
\end{equation}
In practice, Eq.~\eqref{eq:method_normal_equation} is approximately solved using the conjugate gradient method without explicitly forming the matrix inverse. This data-consistency step restricts the reconstruction to the measurement-compatible solution space and mitigates projection-inconsistent structures introduced by the generative prior.

The complete procedure of Lucid is summarized in Algorithm~\ref{alg:method_fmct}.

\begin{algorithm}[t]
\caption{Lucid: Sparsity-Guided Adaptive Flow Matching for Sparse-View CT Reconstruction}
\label{alg:method_fmct}
\begin{algorithmic}[1]
\REQUIRE Sparse-view projection data $\boldsymbol{y}_{\Omega}$; sparse-view system operator $\boldsymbol{A}_{\Omega}$; sparse-view FBP operator $\operatorname{FBP}_{\Omega}$; pretrained velocity network $\boldsymbol{v}_{\boldsymbol{\theta}}$; full-view and sparse-view angle sets $\Theta$ and $\Omega$; number of inference steps $K$; parameters $\lambda_{\mathrm{dc}}$, $\Delta t_{\min}$, $\Delta t_{\max}$, $\alpha$, and $\xi$
\ENSURE Reconstructed image $\boldsymbol{x}^{(K)}$
\STATE Compute the sparsity level $\eta \leftarrow 1-|\Omega|/|\Theta|$
\STATE Compute $\boldsymbol{x}_{\mathrm{FBP}} \leftarrow \operatorname{FBP}_{\Omega}(\boldsymbol{y}_{\Omega})$ and sample $\boldsymbol{z}\sim\mathcal{N}(\boldsymbol{0},\boldsymbol{I})$
\STATE Initialize $\boldsymbol{x}^{(0)} \leftarrow \eta\boldsymbol{z}+(1-\eta)\boldsymbol{x}_{\mathrm{FBP}}$
\STATE Compute the sparsity modulation factor $g(\eta)\leftarrow(1+\alpha\eta)/(1+\alpha)$
\FOR{$k=0$ to $K-1$}
\STATE Set the time condition $t_k \leftarrow \eta(1-k/K)$
\STATE Compute the adaptive step size $\Delta t_k \leftarrow \Delta t_{\min}+(\Delta t_{\max}-\Delta t_{\min})t_k^{\xi}g(\eta)$
\STATE Perform the FM prior update $\tilde{\boldsymbol{x}}^{(k+1)} \leftarrow \boldsymbol{x}^{(k)}-\Delta t_k\boldsymbol{v}_{\boldsymbol{\theta}}(\boldsymbol{x}^{(k)},t_k)$
\STATE Update $\boldsymbol{x}^{(k+1)}$ by approximately solving Eq.~\eqref{eq:method_dc} using CG
\ENDFOR
\STATE \textbf{return} $\boldsymbol{x}^{(K)}$
\end{algorithmic}
\end{algorithm}

\section{Experiments}
\label{sec:experiments}

\begin{table*}[t]
\centering
\caption{Quantitative comparison on the AAPM dataset under different sparse-view settings. Average PSNR (dB) $\uparrow$ and SSIM (\%) $\uparrow$ are reported. The best and second-best results are highlighted in bold and underlined, respectively.}
\label{tab:quantitative_comparison}
\footnotesize
\setlength{\tabcolsep}{5.2pt}
\renewcommand{\arraystretch}{1.15}
\begin{tabular}{llcccccccc}
\toprule
\multirow{2}{*}{Method} & \multirow{2}{*}{Source} & \multicolumn{2}{c}{40 views} & \multicolumn{2}{c}{60 views} & \multicolumn{2}{c}{80 views} & \multicolumn{2}{c}{Average} \\
\cmidrule(lr){3-4} \cmidrule(lr){5-6} \cmidrule(lr){7-8} \cmidrule(lr){9-10}
& & PSNR & SSIM & PSNR & SSIM & PSNR & SSIM & PSNR & SSIM \\
\midrule
FBP         & Baseline & 26.1467 & 47.6630 & 29.4454 & 62.2933 & 32.0325 & 73.1345 & 29.2082 & 61.0303 \\
RGIRT       & PMB~2024  & 32.1878 & 80.7497 & 33.3117 & 84.0287 & 34.3364 & 87.1493 & 33.2786 & 83.9759 \\
STV         & PMB~2025  & 31.2213 & 67.3971 & 33.1710 & 74.2128 & 34.7891 & 79.8726 & 33.0605 & 73.8274 \\
MetaInv-Net & TMI~2020  & 36.2784 & 88.1483 & 36.8868 & 88.9526 & 37.5799 & 90.0631 & 36.9151 & 89.0546 \\
LipCT       & TMI~2026  & \underline{37.7456} & \underline{90.4377} & \underline{38.7311} & \underline{91.2570} & 39.2531 & 91.7659 & 38.5766 & \underline{91.1536} \\
DPS         & ICLR~2023 & 32.2532 & 80.9761 & 32.3100 & 81.0850 & 32.3087 & 81.1056 & 32.2906 & 81.0556 \\
SWORD       & TMI~2024  & 37.6420 & 89.4003 & 38.4232 & 90.6348 & \underline{40.0475} & \underline{92.1135} & \underline{38.7043} & 90.7162 \\
Lucid       & Ours     & \textbf{39.6465} & \textbf{92.4435} & \textbf{40.3130} & \textbf{93.1746} & \textbf{40.7472} & \textbf{93.6818} & \textbf{40.2356} & \textbf{93.0999} \\
\bottomrule
\end{tabular}
\end{table*}

\begin{figure*}[ht]
\centering
\includegraphics[width=\textwidth]{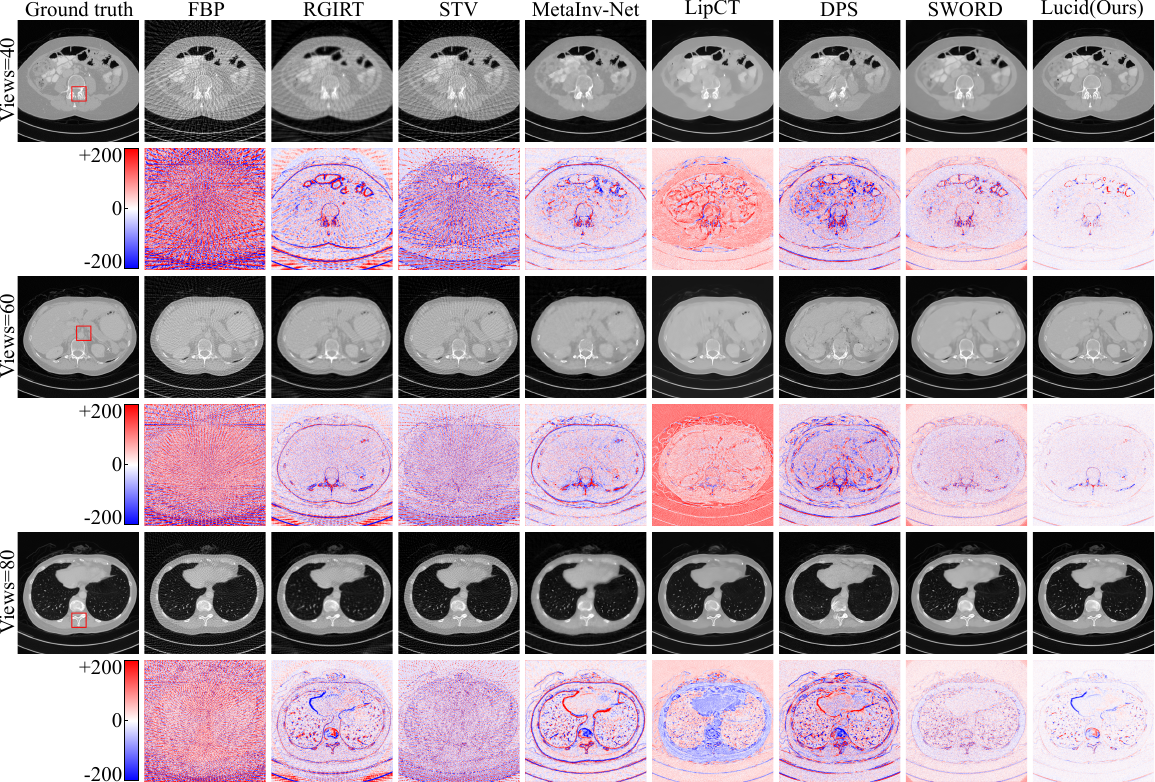}
\caption{Visual comparison and residual maps under \(40\), \(60\), and \(80\) views. For each setting, reconstructed images are displayed with a window of \([-1000,600]\) HU, and residual maps computed as \(\hat{\boldsymbol{x}}-\boldsymbol{x}_{\mathrm{ref}}\) are displayed with a window of \([-200,200]\) HU. Red rectangles mark the ROIs for local comparison.}
\label{fig:visual_comparison}
\end{figure*}

\begin{figure*}[ht]
\centering
\includegraphics[width=\textwidth]{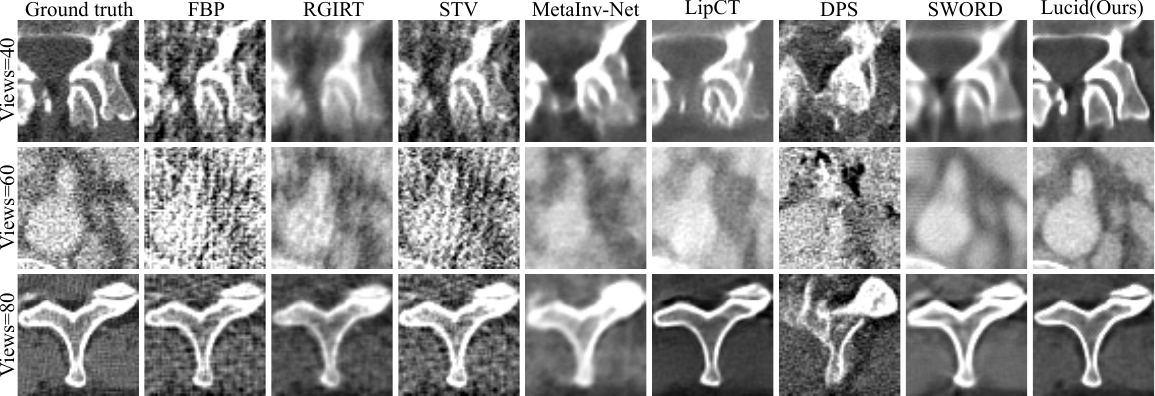}
\caption{Zoomed ROI comparison under \(40\), \(60\), and \(80\) views. The display windows are \([-200,600]\) HU, \([-200,200]\) HU, and \([-200,450]\) HU for \(40\), \(60\), and \(80\) views, respectively.}
\label{fig:visual_comparison_roi}
\end{figure*}

\subsection{Experimental Setup}

\subsubsection{Datasets}

Experiments were conducted on the public 2016 NIH-AAPM-Mayo Clinic Low-Dose CT Grand Challenge dataset~\cite{mccollough2017low}. We selected 1000 normal-dose CT slices from cases L067 and L096 for training and 500 slices from an independent case, L286, for testing, ensuring patient-level separation. All images have a matrix size of \(512 \times 512\). The training images were used only to learn the Flow Matching image prior, while full-view reconstructions from 720 uniformly sampled projections were used as reference images for testing. All competing methods were evaluated on the same test slices and projection data.

A two-dimensional fan-beam CT geometry was simulated using the ASTRA Toolbox. The source-to-isocenter distance and source-to-detector distance were \(540~\mathrm{mm}\) and \(950~\mathrm{mm}\), respectively. The detector contained 900 bins with a bin size of \(1.1~\mathrm{mm}\). Full-view sinograms were generated using 720 uniformly spaced views over \(360^{\circ}\), and sparse-view measurements were obtained by uniformly selecting \(40\), \(60\), or \(80\) views. No additional Poisson or electronic noise was added, so that the experiments focused on angular undersampling.

\subsubsection{Training Details}

The proposed method was implemented in PyTorch, and all experiments were performed on an NVIDIA GeForce RTX 4090 GPU. The Flow Matching prior used a two-dimensional U-Net to estimate the velocity field, with one input and one output channel. The network was trained only on normal-dose CT images using AdamW with an initial learning rate of \(1\times10^{-4}\), a batch size of 1, and 200 epochs. The continuous time variable was discretized into 1000 levels during training.

During inference, the same pretrained model was used for all sparse-view settings. We fixed \(K=50\), \(\Delta t_{\min}=0.006\), \(\Delta t_{\max}=0.09\), \(\alpha=0.99\), \(\xi=1\), and \(\lambda_{\mathrm{dc}}=0.9\). The data-consistency subproblem was solved by the conjugate gradient method.

\subsection{Comparison with Existing SVCT Methods}

We compared Lucid with representative analytical, model-based iterative, learning-based, and generative prior-based SVCT methods, including FBP, RGIRT~\cite{zhang2024robust}, STV~\cite{yu2025springback}, MetaInv-Net~\cite{zhang2020metainv}, LipCT~\cite{shi2025prompting}, DPS~\cite{chung2022diffusion}, and SWORD~\cite{xu2024stage}.

PSNR and SSIM were used for quantitative evaluation. As shown in Table~\ref{tab:quantitative_comparison}, Lucid achieves the best performance under all three sparse-view settings, with average PSNR and SSIM values of \(40.2356\) dB and \(93.0999\%\), respectively. Compared with the second-best methods, Lucid improves the average PSNR by \(1.5313\) dB over SWORD and the average SSIM by \(1.9463\) percentage points over LipCT.

Figs.~\ref{fig:visual_comparison} and~\ref{fig:visual_comparison_roi} present the full-image comparison, residual maps, and zoomed ROIs under \(40\), \(60\), and \(80\) views. In the full images, FBP exhibits severe streak artifacts, especially at \(40\) views, while model-based iterative methods reduce part of the artifacts but still suffer from residual streaks and structural blurring. Learning-based methods further improve visual quality, but they may smooth anatomical boundaries or weaken low-contrast structures. Diffusion-based methods suppress global artifacts more strongly, yet local structural fluctuations remain visible in some regions. In contrast, Lucid produces cleaner reconstructions across all view settings, with the improvement being particularly pronounced under the challenging \(40\)-view case.

The residual maps provide a clearer comparison of reconstruction errors. FBP and iterative methods show strong global residual patterns caused by angular undersampling, whereas learning-based and diffusion-based methods reduce these errors but still present structured residuals near high-contrast edges and anatomical boundaries. Lucid yields weaker and more spatially localized residuals, indicating better consistency with the reference images. The zoomed ROIs further show that Lucid better preserves local anatomical details, including soft-tissue boundaries and high-contrast structures, while competing methods either leave residual artifacts or introduce excessive smoothing. These visual results are consistent with the quantitative results and demonstrate that Lucid achieves a favorable balance between artifact suppression, structural fidelity, and data consistency.

\subsection{Generalization Across Different Sparse-View Settings}

\begin{figure}[htbp]
\centering
\includegraphics[width=0.48\textwidth]{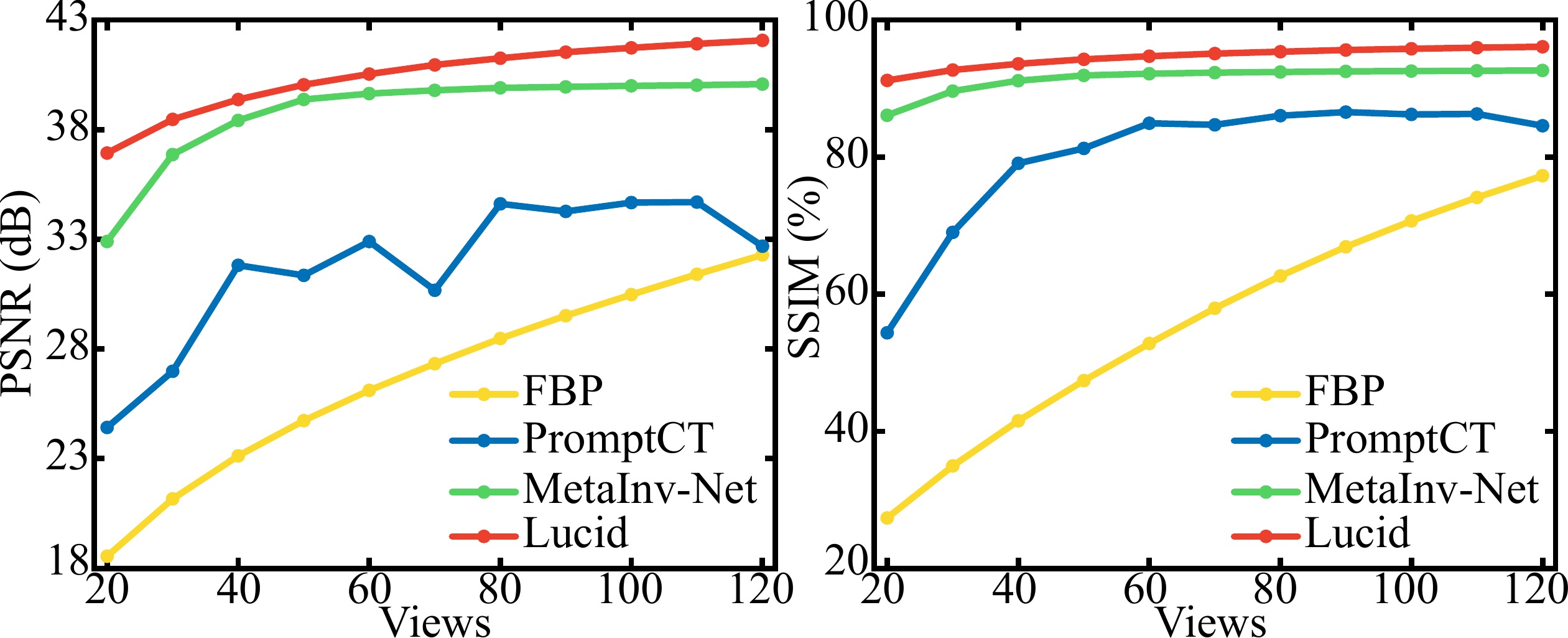}
\caption{Reconstruction performance under varying sampling densities from \(20\) to \(120\) views with an interval of \(10\). Lucid uses the same pretrained Flow Matching prior for all settings.}
\label{fig:generalization_across_views}
\end{figure}

This subsection evaluates the adaptability of Lucid to continuously varying sampling densities. We varied the number of projection views from \(20\) to \(120\) and compared Lucid with FBP, PromptCT~\cite{shi2025prompting}, and MetaInv-Net~\cite{zhang2020metainv}. For all view settings, Lucid used the same pretrained Flow Matching prior without view-dependent retraining.

As shown in Fig.~\ref{fig:generalization_across_views}, FBP improves as more views are available but remains limited by the lack of an image prior. PromptCT generally outperforms FBP, but its performance fluctuates at intermediate view numbers, possibly because discrete prompt representations may not monotonically describe continuously varying sampling densities. MetaInv-Net shows a more stable trend due to its unfolding structure and explicit imaging model.

Lucid achieves the highest PSNR and SSIM across the entire \(20\)--\(120\)-view range and exhibits a smooth improvement as the number of views increases. These results indicate that the proposed sparsity-aware initialization, adaptive step-size control, and projection-domain data consistency enable Lucid to generalize across different sparse-view settings using a single pretrained model.

\subsection{Analysis of Structural Fidelity and Hallucination-Related Errors}

\begin{figure}[t]
\centering
\includegraphics[width=0.48\textwidth]{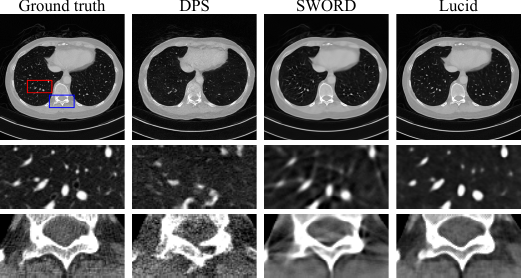}
\caption{Visual comparison of generative reconstruction methods for structural fidelity and hallucination-related errors. The columns show the reference image, DPS, SWORD, and Lucid. The rows show the full image, lung-parenchyma ROI, and bone ROI, with display windows of \([-1000,600]\) HU, \([-1000,120]\) HU, and \([-200,410]\) HU, respectively.}
\label{fig:hallucination_analysis}
\end{figure}

\begin{table}[htbp]
\centering
\caption{ROI-based evaluation of hallucination-related local errors. Lower values indicate better agreement with the reference image.}
\label{tab:hallucination_roi_metrics}
\begin{tabular}{ccccc}
\toprule
\multirow{2}{*}{Method} 
& \multicolumn{2}{c}{Lung ROI} 
& \multicolumn{2}{c}{Bone ROI} \\
\cmidrule(lr){2-3} \cmidrule(lr){4-5}
& MAE & HFEN & MAE & HFEN \\
\midrule
DPS 
& 0.0163 & 1.3246 
& 0.0184 & 1.2380 \\
SWORD 
& 0.0079 & 0.6188 
& 0.0106 & 0.7493 \\
Lucid 
& \textbf{0.0074} & \textbf{0.5253} 
& \textbf{0.0088} & \textbf{0.5817} \\
\bottomrule
\end{tabular}
\end{table}

Generative priors can suppress sparse-view artifacts, but may also introduce hallucination-like local structures under severe angular undersampling. We therefore compare Lucid with DPS and SWORD in terms of local structural fidelity. As shown in Fig.~\ref{fig:hallucination_analysis}, DPS introduces unstable local textures in the lung and bone regions, while SWORD reduces these perturbations but tends to over-smooth or distort fine structures. In contrast, Lucid better preserves lung parenchymal structures and bony contours, producing local details more consistent with the reference image.

Table~\ref{tab:hallucination_roi_metrics} further reports ROI-based MAE and high-frequency error norm (HFEN)~\cite{ravishankar2011mr}. MAE evaluates local intensity fidelity, while HFEN emphasizes high-frequency structural discrepancies after Laplacian-of-Gaussian filtering. Lucid achieves the lowest MAE and HFEN in both ROIs, indicating fewer spurious high-frequency structures and better local agreement with the reference image. These results suggest that the combination of sparsity-aware Flow Matching inference and projection-domain data consistency reduces hallucination-like structural errors while preserving the artifact-suppression capability of generative reconstruction.

\subsection{Ablation Studies}

\begin{table*}[t]
\centering
\caption{Ablation results of sparsity-guided initialization (SGI) and adaptive step-size scheduling (AS). Average PSNR (dB) $\uparrow$ / SSIM (\%) $\uparrow$ are reported, with the best results highlighted in bold.}
\label{tab:ablation_sgi_as}
\footnotesize
\setlength{\tabcolsep}{8pt}
\renewcommand{\arraystretch}{1.15}
\begin{tabular}{l>{\centering\arraybackslash}p{0.035\textwidth}>{\centering\arraybackslash}p{0.035\textwidth}cccc}
\toprule
Variant & SGI & AS & 40 views & 60 views & 80 views & Average \\
\midrule
w/o SGI \& AS & $\times$ & $\times$ & 30.5373 / 77.5303 & 31.7467 / 84.2614 & 31.9497 / 84.2660 & 31.4112 / 82.0192 \\
w/o AS        & $\checkmark$ & $\times$ & 30.7090 / 85.6633 & 31.3396 / 87.7036 & 31.5703 / 88.0883 & 31.2063 / 87.1517 \\
w/o SGI       & $\times$ & $\checkmark$ & 31.2081 / 80.6547 & 32.1922 / 84.5899 & 32.4555 / 85.0571 & 31.9519 / 83.4339 \\
Lucid         & $\checkmark$ & $\checkmark$ & \textbf{39.6465} / \textbf{92.4435} & \textbf{40.3130} / \textbf{93.1746} & \textbf{40.7472} / \textbf{93.6818} & \textbf{40.2356} / \textbf{93.0999} \\
\bottomrule
\end{tabular}
\end{table*}

\begin{figure}[t]
\centering
\includegraphics[width=0.48\textwidth]{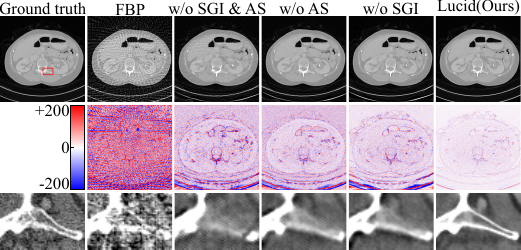}
\caption{Visual ablation comparison under the \(40\)-view setting. The columns show the reference image, FBP initialization, w/o SGI \& AS, w/o AS, w/o SGI, and Lucid. The rows show reconstructed images, residual maps, and zoomed ROIs marked by the red rectangle, with display windows of \([-1000,600]\) HU, \([-200,200]\) HU, and \([-200,400]\) HU, respectively.}
\label{fig:ablation_visual_40views}
\end{figure}

To evaluate the contributions of sparsity-guided initialization (SGI) and adaptive step-size scheduling (AS), we compared four variants under \(40\), \(60\), and \(80\) views: w/o SGI \& AS, w/o AS, w/o SGI, and the full Lucid model. All other settings were kept unchanged.

As shown in Table~\ref{tab:ablation_sgi_as}, using either SGI or AS alone provides only partial improvement. SGI mainly improves structural similarity, whereas AS provides a more stable gain in reconstruction accuracy. When both components are used, Lucid achieves the best PSNR and SSIM under all sparse-view settings, demonstrating their complementarity.

Fig.~\ref{fig:ablation_visual_40views} further confirms this observation under the challenging \(40\)-view setting. Removing either component leaves visible residual artifacts, boundary blurring, or loss of fine details. In contrast, the full Lucid model produces weaker residuals and better preserves local high-contrast structures, indicating improved artifact suppression and structural fidelity.

\subsection{Sensitivity Analysis of Step-Size Parameters}

\begin{figure}[t]
  \centering
  \includegraphics[width=0.48\textwidth]{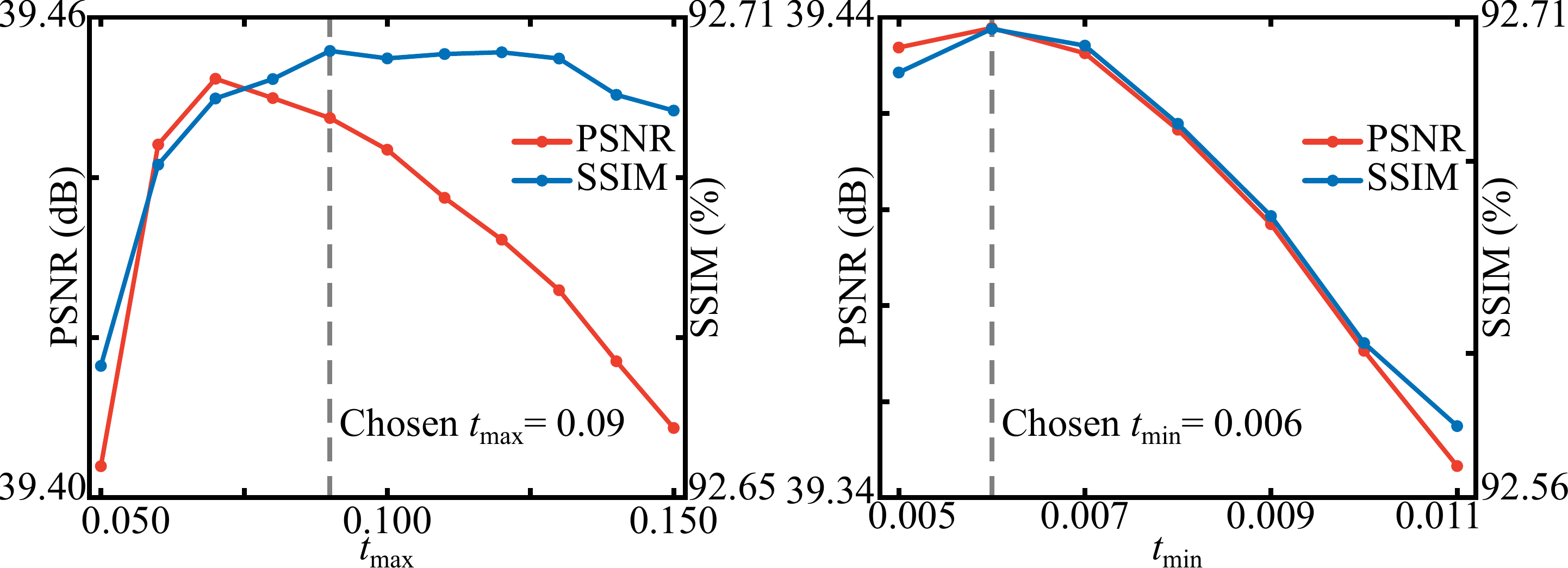}
  \caption{Sensitivity analysis of the step-size bounds \(\Delta t_{\min}\) and \(\Delta t_{\max}\) under the \(40\)-view setting.}
  \label{fig:parameter_sensitivity}
\end{figure}

We evaluated the sensitivity of Lucid to the step-size bounds \(\Delta t_{\min}\) and \(\Delta t_{\max}\) under the \(40\)-view setting using 50 randomly selected test slices. As shown in Fig.~\ref{fig:parameter_sensitivity}, fixing \(\Delta t_{\min}=0.006\) and varying \(\Delta t_{\max}\) from \(0.05\) to \(0.15\) leads to only mild changes in PSNR and SSIM, with \(\Delta t_{\max}=0.09\) providing a favorable trade-off. Similarly, when fixing \(\Delta t_{\max}=0.09\) and varying \(\Delta t_{\min}\) from \(0.005\) to \(0.011\), the performance remains stable and peaks around \(\Delta t_{\min}=0.006\). We therefore set \(\Delta t_{\min}=0.006\) and \(\Delta t_{\max}=0.09\) in all experiments. These results indicate that Lucid is not sensitive to the precise choice of the step-size bounds and does not require delicate tuning of these parameters.

\section{Conclusion}
\label{sec:conclusion}

In this paper, we proposed Lucid, a sparsity-adaptive and data-consistent reconstruction framework for sparse-view CT based on deterministic Flow Matching. Lucid learns a generative image prior from high-quality CT images without relying on view-specific supervised training. During inference, the sampling sparsity level is explicitly incorporated to guide the initialization and step-size scheduling, while projection-domain data consistency constrains the reconstruction with the measured projections. This design enables the pretrained prior to adapt to different degrees of angular undersampling within a unified reconstruction process.

Experimental results show that Lucid achieves superior quantitative and visual performance over representative model-based, learning-based, and generative reconstruction methods under multiple sparse-view settings. It also generalizes to continuously varying numbers of projection views using the same pretrained model. In addition, ROI-based analyses demonstrate that Lucid better preserves lung parenchymal structures, bony edges, and low-contrast soft-tissue details, while reducing hallucination-like local structures observed in diffusion-based generative reconstruction. Ablation studies further confirm the complementary roles of sparsity-guided initialization and adaptive step-size scheduling.

Overall, Lucid provides an effective and interpretable generative reconstruction strategy for sparse-view CT with varying sampling densities. Future work will extend the framework to more realistic acquisition scenarios, including noisy projections, cross-scanner protocols, and three-dimensional cone-beam CT, and will investigate more efficient data-consistency solvers to reduce computational cost.

\bibliographystyle{IEEEtran}
\bibliography{Bibliography}

\newpage

\vspace{11pt}

\vfill

\end{document}